%% file: main.tex
\documentclass[10pt,twocolumn,letterpaper]{article}

\usepackage{cvpr}              
\usepackage{multirow}
\usepackage[inkscapelatex=false]{svg}
\usepackage{bm}
\usepackage{subcaption}
\usepackage{graphicx}

\newcommand{\parsection}[1]{\noindent\textbf{#1:}~}

\definecolor{cvprblue}{rgb}{0.21,0.49,0.74}
\usepackage[pagebackref,breaklinks,colorlinks,citecolor=cvprblue]{hyperref}
\usepackage[accsupp]{axessibility}
\def\paperID{9221} 
\def\confName{CVPR}
\def\confYear{2024}

\title{CAT-DM: Controllable Accelerated Virtual Try-on with Diffusion Model}

\author{
{Jianhao Zeng{$^{1}$}} \qquad Dan Song{$^{1\ast}$}  \qquad Weizhi Nie{$^{1}$} \qquad  Hongshuo Tian{$^{1}$}\\
\qquad Tongtong Wang{$^{2}$} \qquad An-An Liu{$^{1\ast}$} \\
\normalsize
$^{1}$\    Tianjin University ~~ $^{2}$\,Tencent LightSpeed Studio\\
}
\begin{document}

\twocolumn[{%
\renewcommand\twocolumn[1][]{#1}%
\maketitle
    \includegraphics[width=1\linewidth]{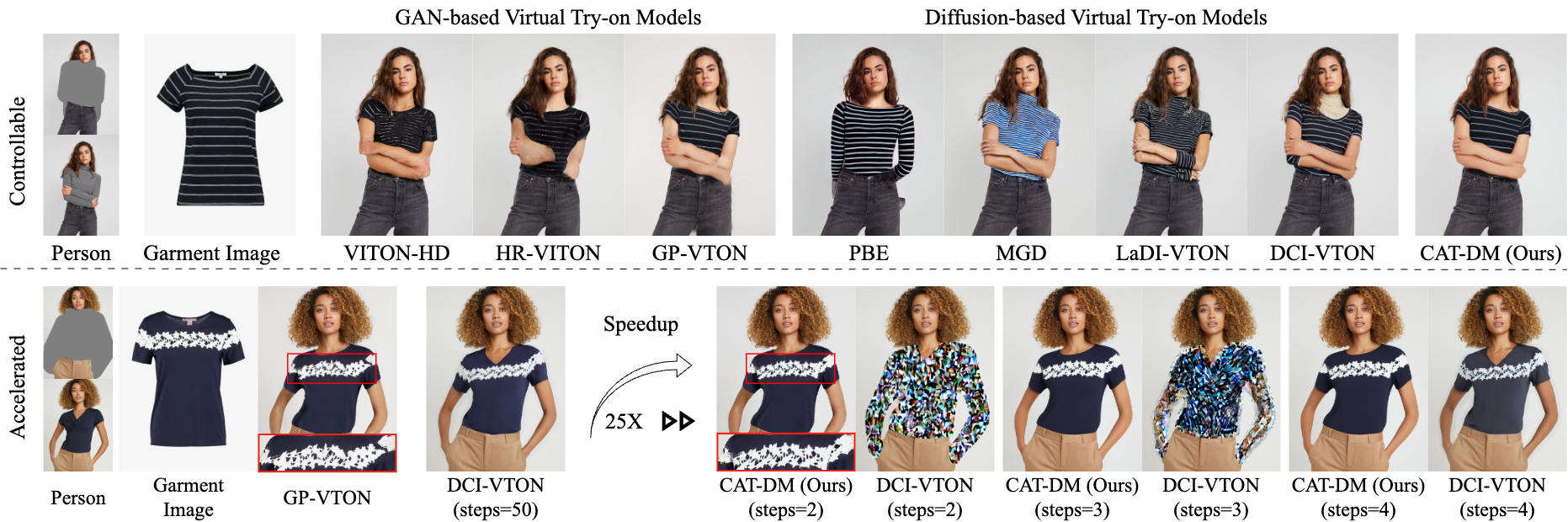}
    \captionof{figure}{
CAT-DM not only enhances the controllability of the image generation process for virtual try-on but also effectively accelerates the sampling speed of the diffusion models.
\textbf{Top:} Comparison results with other methods. 
CAT-DM accurately generates the pattern details on garments and produces images that are sufficiently clear.
\textbf{Bottom:} CAT-DM requires fewer sampling steps than other diffusion models to generate clear and realistic virtual try-on images. Compared to the default 50 sampling steps of DCI-VTON~\cite{DCI-VTON}, CAT-DM achieves a 25-fold acceleration.
}
    \vspace{3mm}
    \label{Fig:1_Teaser}
}]

\input{sec/0_abstract}    
\input{sec/1_introduction}
\input{sec/2_related_work}

\input{sec/3_Method}

\input{sec/4_Experiments}

\input{sec/5_Conclusion}

{
    \small
    \bibliographystyle{ieeenat_fullname}
    \bibliography{main}
}


\end{document}

%% file: sec/0_abstract.tex
\begin{abstract}
\renewcommand{\thefootnote}{\fnsymbol{footnote}}
\footnotetext[1]{Corresponding Author: Dan Song (dan.song@tju.edu.cn), An-An Liu (anan0422@gmail.com). The code is available at \url{https://github.com/zengjianhao/CAT-DM}.}
\footnotetext[2]{The research is supported in part by the National Natural Science Foundation of China (U21B2024, 62232337, 61902277).}

Generative Adversarial Networks (GANs) dominate the research field in image-based virtual try-on, but have not resolved problems such as unnatural deformation of garments and the blurry generation quality.
While the generative quality of diffusion models is impressive, achieving controllability poses a significant challenge when applying it to virtual try-on and multiple denoising iterations limit its potential for real-time applications.
In this paper, we propose \textbf{C}ontrollable \textbf{A}ccelerated virtual \textbf{T}ry-on with \textbf{D}iffusion \textbf{M}odel (CAT-DM).
To enhance the controllability, a basic diffusion-based virtual try-on network is designed, which utilizes ControlNet to introduce additional control conditions and improves the feature extraction of garment images.
In terms of acceleration, CAT-DM initiates a reverse denoising process with an implicit distribution generated by a pre-trained GAN-based model. 
Compared with previous try-on methods based on diffusion models, CAT-DM not only retains the pattern and texture details of the in-shop garment but also reduces the sampling steps without compromising generation quality.
Extensive experiments demonstrate the superiority of CAT-DM against both GAN-based and diffusion-based methods in producing more realistic images and accurately reproducing garment patterns.

\end{abstract}

%% file: sec/1_introduction.tex
\section{Introduction}

Image-based virtual try-on has emerged as a prominent and popular research topic within the field of AIGC, particularly focusing on conditional person image generation. By taking a person image and a target garment image as inputs, the objective of this task is to generate a photo of the person seamlessly wearing the desired garment.
Requirements are expected in both aspects of person and clothes: 1) the posture and identity such as face and skin should be the same as the person; 2) target garment is naturally warped and seamlessly put on the body without losing characteristics such as pattern and texture.
 
Existing image-based virtual try-on methods~\cite{PFAFN, ClothFlow, Viton, StyleGAN, CP-VTON+, CP-VTON, ACGPN, VITON-HD, HR-VITON, GP-VTON} primarily rely on Generative Adversarial Networks (GANs)~\cite{GANs}. 
GAN-based try-on methods typically start by warping the in-shop garment image to match the given person image, then combine the warped image with the person image into a generator for synthesis.
However, existing garment deformation  approaches such as TPS~\cite{TPS}, STN~\cite{STN} and FlowNet~\cite{FlowNet} are not flexible enough to deal with challeging poses (\cref{Fig:1_Teaser}).
Additionally, images produced by GAN-based methods often lack a degree of realism and may fail to generate finer details.

Recently, diffusion models have garnered widespread applications in the field of image generation, demonstrating outstanding performance across various tasks such as super-resolution~\cite{SRDiff}, image restoration~\cite{DDRM} and text-guided image generation~\cite{DreamBooth, Imagen}. 
Compared with GANs, diffusion models demonstrate enhanced stability~\cite{ModeCollapse} during training and excel in producing images with fine-grained realism such as clearer hands and arms.
However, when applying diffusion models to virtual try-on task \cite{PBE,MGD,LaDI-VTON,DCI-VTON}, the controllability of the generated results, particularly preserving complex textures and patterns of target garment, remains challenging (\cref{Fig:1_Teaser}).
Furthermore, the generation of high-fidelity images via diffusion models requires a considerable number of sampling steps, thereby limiting their application in real-time virtual try-on scenarios.

To enhance the controllability of diffusion models, we propose a Garment-Conditioned Diffusion Model (GC-DM). This model utilizes the ControlNet~\cite{ControlNet} architecture to provide more garment-agnostic person representations as control conditions. On the other hand, GC-DM improves the feature extraction of garment images, providing the model with more detailed garment information to better control the pattern generation in the try-on images. In addition, to ensure that the image areas outside of the garment region remain unchanged, GC-DM employs Poisson blending~\cite{Poisson} to seamlessly integrate the original person images with the generated try-on images.

Building upon the foundation of GC-DM, we further propose a truncation-based acceleration strategy to accelerate the inference speed of diffusion models.
Inspired by truncated diffusion probabilistic model (TDPM)~\cite{TDPM}, we employ an implicit distribution to provide initial samples for the reverse denoising process, instead of using the Gaussian noise as the starting point for reverse denoising, thereby significantly reducing the sampling steps. 
Unlike TDPM which learns an implicit distribution, we utilize a pre-trained GAN-based model to generate an initial try-on image, and then add noise to this image to obtain the implicit distribution.

In summary, the proposed CAT-DM is equipped with GC-DM, a new diffusion-based virtual try-on model, and a truncation-based acceleration strategy initialized by the coarse image generated by a pre-trained GAN.
As shown in \cref{Fig:1_Teaser}, CAT-DM capitalizes on the robust generative capabilities of diffusion models, as well as the controllability of GAN-based models, meanwhile significantly reducing the number of required sampling steps. Extensive experiments validate the effectiveness of each component of CAT-DM, which achieves state-of-the-art results on two widely used benchmarks~\cite{VITON-HD, DressCode}.

The main contributions of our work are: 
(1) We propose CAT-DM, a virtual try-on model, to create high-fidelity images using fewer sampling steps.
(2) We propose GC-DM to improve the controllability of diffusion models by offering additional control conditions and enhancing the extraction of garment image features.
(3) We introduce a truncation-based acceleration strategy to synthesize the advantages of both GAN-based models and diffusion models, and reduce the sampling steps.

%% file: sec/2_related_work.tex
\section{Related Work}

\subsection{Image-based Virtual Try-On}
A recent survey~\cite{survey} has comprehensively reviewed previous SOTA methods in image-based virtual try-on.
GAN-based models play a dominant role in this research area, which usually warp target clothes first and then synthesize try-on images conditioned with warped clothes and person image. 
Different strategies have been tried for clothing warping, such as thin-plate-spline (TPS) interpolation~\cite{Viton}, spatial transformation network (STN)~\cite{OVNet} and flow estimation~\cite{GP-VTON}. However, as shown in \cref{Fig:1_Teaser}, complex or unconventional postures are still challenging for GAN-based methods.
In terms of generation quality, some high-resolution methods are explored, e.g., VITON-HD~\cite{VITON-HD} introduces a misalignment-aware normalization and HR-VTON~\cite{HR-VITON} further refines it by predicting both segmentation and flow simultaneously. During the training process, GANs~\cite{GANs} usually encounter issues such as mode collapse~\cite{ModeCollapse}.

Diffusion models significantly improve the realism of generated images, and several approaches~\cite{TryOnDiffusion,Linking,PoseGarment,Size} apply this advanced model to image-based virtual try-on. PBE~\cite{PBE} is a robust diffusion model for image generation, capable of semantically altering image content based on exemplar image.
MGD~\cite{MGD} uses the multimodal data to guide the generation of fashion images. However, as shown in \cref{Fig:1_Teaser}, the controllability of generated contents is not well addressed. 
Later LaDI-VTON~\cite{LaDI-VTON} uses a textual inversion component to map the visual features of garments to the CLIP token embedding space.
DCI-VTON~\cite{DCI-VTON} pastes the warped garment to the input of the diffusion model as the local condition to better retain the characteristics of the garments. Although the controllability is improved, diffusion-based methods still suffer from redundant sampling steps.
Therefore, in this paper we propose a controllable and accelerated model to both enhance the generation quality and speed.

\subsection{Diffusion Models}

Diffusion models are a class of generative models that learn the target distribution through an iterative denoising process. Compared to GANs, diffusion models can generate images with more intricate details, resulting in higher quality outputs. Denoising diffusion probabilistic models (DDPMs)~\cite{DDPM} consist of a Markovian forward process that gradually corrupts the data sample $\mathbf{x}_0$ into the Gaussian noise $\mathbf{x}_T$, and a learnable reverse process that converts $\mathbf{x}_T$ back to $\mathbf{x}_0$ iteratively. 
Assuming the parameter of the diffusion model is denoted as $\theta$, the training process of the diffusion model can be described as:

\begin{equation}
    \mathbb{E}_{\mathbf{x}_0,t,\epsilon}\left[\left \Vert \epsilon-\epsilon_\theta(\sqrt{\bar{\alpha}_t}\mathbf{x}_0+\sqrt{1-\bar{\alpha}_t}\epsilon,t) \right \Vert^2_2\right],
    \label{eq1}
\end{equation}
where $\mathbf{x}_0$ is from the truth data distribution, $t$ follows a uniform distribution over the set $\{1,2,...,T\}$, $\epsilon\sim\mathcal{N}(0,\mathbf{I})$ represents randomly generated noise and $\bar{\alpha}_t=\prod^t_{s=1}\alpha_s$ is a pre-defined variance schedule in $t$ steps. Therefore, $\sqrt{\bar{\alpha}_t}\mathbf{x}_0+\sqrt{1-\bar{\alpha}_t}\epsilon$ represents the noisy image after adding noise to $\mathbf{x}_0$. The diffusion model is trained by predicting the added noise $\epsilon$ from the noisy image.

After the diffusion model is trained, diffusion models can be used to synthesize new images by taking a random noise sample  $\mathbf{x}_T\sim\mathcal{N}(0,\mathbf{I})$ and  denoise it for $1\le t\le T$ iteratively:

\begin{equation}
    \mathbf{x}_{t-1}=\frac{1}{\sqrt{\alpha_t}}\left (\mathbf{x}_t-\frac{1-\alpha_t}{\sqrt{1-\bar{\alpha}_t}}\epsilon_{\theta}(\mathbf{x}_t,t) \right )+\sigma_t \mathbf{z},
    \label{eq2}
\end{equation}
where $\mathbf{z}\sim\mathcal{N}(0,\mathbf{I})$ and $\sigma_t^2=\frac{1-\bar{\alpha}_{t-1}}{1-\bar{\alpha}_t}(1-\alpha_t)$. DDPMs are designed to make the denoising process closely resemble a Gaussian distribution. To achieve this, they often set the parameter T to a large value, typically around 1000. As a result, DDPMs require a sequence of 1000 noise prediction steps for image generation.

\begin{figure*}
    \centering
    \includegraphics[width=1\linewidth]{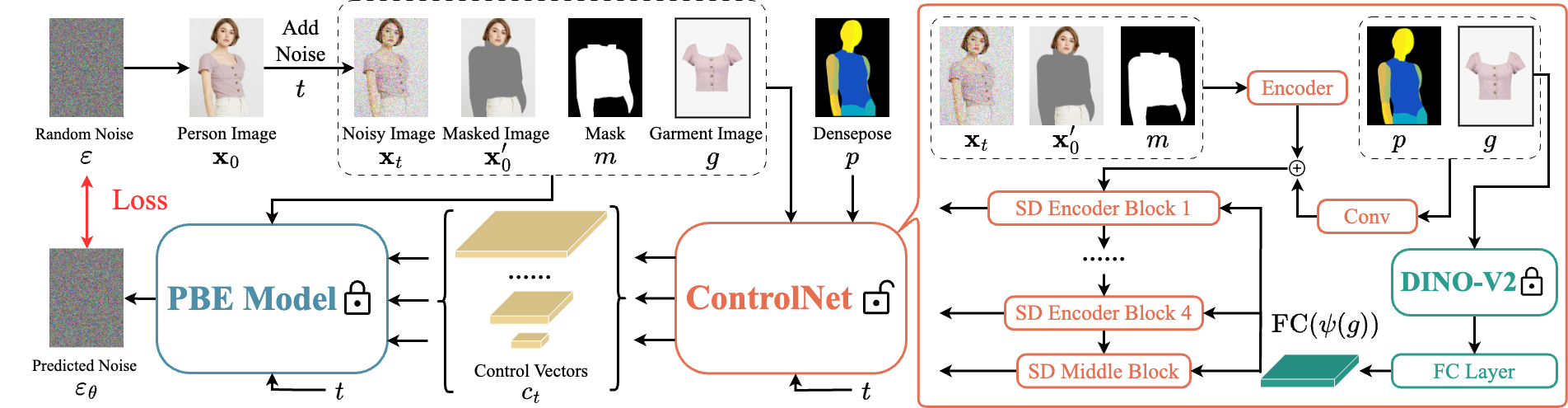}
    \caption{
The training pipeline of the GC-DM in our method. GC-DM comprises a fixed-parameter PBE and a trainable ControlNet. Apart from the given noisy image $\mathbf{x}_t$, time steps $t$, mask $m$, masked image $\mathbf{x}_0'$ and garment image $g$, ControlNet generates a set of control vectors $c_t$ by incorporating additional control conditions, such as densepose $p$. Control vectors are incorporated into the PBE to enhance the model's controllability while preserving the PBE's generative capabilities.
}
    \vspace{-4mm}
    \label{Fig:2_GC-DM}
\end{figure*}

\subsection{Accelerated Sampling}

While diffusion models can produce realistic images, the necessity for multiple sampling steps to generate a single image limits their application in real-time virtual try-on scenarios. Denoising diffusion implicit models (DDIMs)~\cite{DDIM} introduce a non-Markovian diffusion process that accelerates sampling without the need for additional training. This approach has been widely adopted in the field. Progressive distillation~\cite{PD} is a highly effective approach for boosting the sampling rate of a diffusion model through repeated distillation. However, this repeated distillation comes with considerable training expenses. Although consistency model~\cite{Consistency} can achieve accelerated sampling by employing only one distillation, its performance is often inadequate for many scenarios~\cite{Catch}, which hampers its broader practical application. TDPM~\cite{TDPM} shortens the diffusion trajectory by learning an implicit distribution at step $T_{\text{trunc}}$, which initiates the reverse diffusion process. 
However, it is hard to learn the corrupted data at step $T_{\text{trunc}}$. On the other hand, these methods of accelerating sampling inevitably compromise the model's performance to some extent and are unable to resolve the issue of limited controllability in diffusion models.
Unlike directly predicting the implicit distribution at step $T_{\text{trunc}}$, CAT-DM first generates an initial try-on image using a pre-trained GAN-based model, and then obtains the data distribution at step $T_{\text{trunc}}$, by adding noise. Our approach not only achieves accelerated sampling but also enhances the controllability of diffusion models.

%% file: sec/3_Method.tex
\section{Method}

To address the issues faced by diffusion models in virtual try-on tasks, such as the loss of garment pattern details and the necessity for numerous sampling steps, we propose CAT-DM. 
It consists of GC-DM, a novel garment-conditioned diffusion model designed for virtual try-on to enhance the controllability, and the truncation-based acceleration strategy with a pre-trained GAN-based model as initialization for acceleration.

\subsection{Garment-Conditioned Diffusion Model}
\label{subsection:GC-DM}

The GC-DM employs a ControlNet~\cite{ControlNet} architecture, which introduces additional control conditions while preserving the generative capabilities of PBE\cite{PBE}. It enhances the feature extractor to provide more detailed information about garment, thus improving control over the generation of garment areas. For areas outside of garment, GC-DM uses a Poisson blending~\cite{Poisson} to ensure that the original person information remains unchanged. The specific designs will be elaborated below.

\textbf{ControlNet architecture.}
Diffusion models excel in image generation but require substantial computational resources and GPU memory due to a large number of parameters, limiting their application and advancement. Additionally, they mainly provide semantic-level control, and improving this controllability without retraining billion-parameter model is challenging.

As shown in \cref{Fig:2_GC-DM}, GC-DM consists of PBE with locked-parameters and a trainable ControlNet. PBE is a robust image generation model capable of semantically altering image content based on exemplar image, and trained on millions of images. This model utilizes an U-Net~\cite{UNet} architecture, comprising multiple SD Encoder Blocks, several SD Decoder Blocks, and an SD Middle Block. The SD Encoder Blocks and SD Decoder Blocks are interconnected via skip-connections. However, PBE is not directly suited for the virtual try-on task, as this task requires the generated try-on images to remain pixel-consistent with the target garment images. Additionally, since PBE has hundreds of millions of parameters, the training of which requires significant computational resources.

We lock all parameters of PBE and copy the parameters of SD Encoder Blocks and SD Middle Block to ControlNet. During the training process, we perform gradient updates exclusively on the parameters of ControlNet. This approach accelerates training and conserves GPU memory, thereby reducing the diffusion model's demand for computational resources.

Given a person image $\mathbf{x}_0$, GC-DM progressively add noise to the image and produces a noisy image $\mathbf{x}_t$ , with $t$ being how many times the noise is added. Given a set of conditions including noisy image $\mathbf{x}_t$, time steps $t$, mask $m$, masked image $\mathbf{x}_0'$, garment image $g$ as well as additional control conditions (such as densepose $p$), ControlNet generates a set of control vectors $c_t$. These vectors are integrated into the skip-connections and the SD Middle Block of PBE's U-Net architecture, thereby directing the generation process of the PBE. Similar to \cref{eq1}, GC-DM learns a network $\epsilon_\theta$ to predict the noise added to the noisy image $\mathbf{x}_t$ with: 

\begin{equation}
    \mathbb{E}_{\mathbf{x}_0,\mathbf{x}_0', m, g, p,t,\epsilon}\left[\left \Vert \epsilon-\epsilon_\theta(\mathbf{x}_t,\mathbf{x}_0', m,g,p,t) \right \Vert^2_2\right]
    \label{eq3}
\end{equation}

By introducing more garment-agnostic person representations as control conditions through ControlNet, GC-DM can not only preserve the generative capabilities of the PBE but also enhance the controllability of the diffusion model.

\begin{figure*}
    \centering
    \includegraphics[width=1\linewidth]{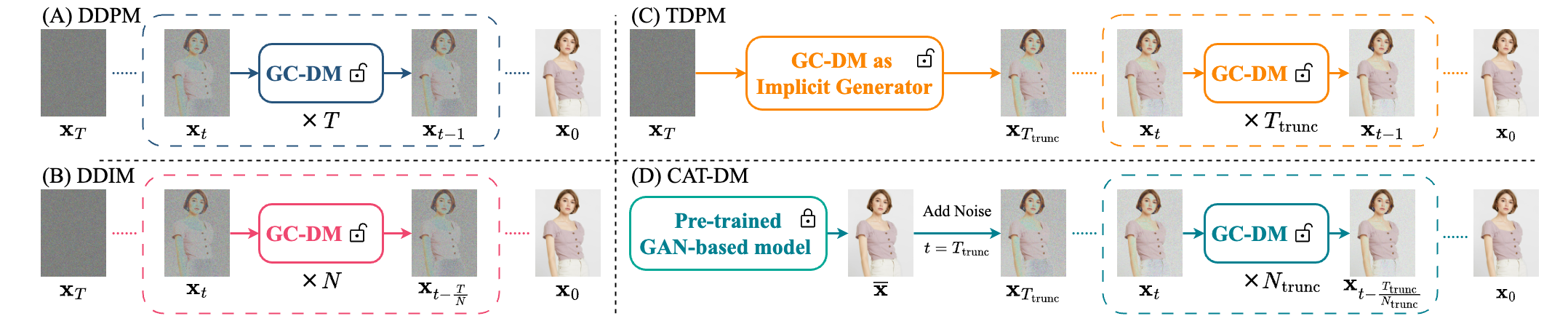}
    \caption{
Illustration of different sampling methods in diffusion models. (A) The conventional DDPMs~\cite{DDPM} denoise gradually with a large number of time steps $T$. (B) DDIMs~\cite{DDIM} employ a class of non-Markovian diffusion processes to denoise gradually. Compared to DDPMs, DDIMs requires fewer sampling steps, that is, $N\ll T$. (C) TDPM~\cite{TDPM} repurposes the parameter of the diffusion model to generate the implicit distribution at step $T_{\text{trunc}}$, using it as the initial sample for the reverse diffusion process. This approach accelerates sampling, resulting in $T_{\text{trunc}}\ll T$. (D) CAT-DM utilizes a pre-trained GAN-based model to generate an initial try-on image $\bar{\mathbf{x}}$, which is then subjected to noise addition, making the noisy image $\mathbf{x}_{T_{\text{trunc}}}$ as the starting point of the reverse diffusion process.}

    \label{Fig:3_CAT-DM}
\end{figure*}

\textbf{Garment feature extraction.} 
Although PBE is capable of generating realistic images based on example images, its lack of pixel-level control often results in inaccurate reconstruction of patterns on garment images in the virtual try-on task ( as shown in \cref{Fig:1_Teaser}). The underlying issue stems from PBE's use of CLIP~\cite{CLIP} as an encoder $\psi$ to extract feature information from garment images. While CLIP can align image information with corresponding textual descriptions in a shared space, it falls short in the virtual try-on task. The semantic information extracted by CLIP is insufficient to accurately describe the patterns and texture details of garment images.

To grant GC-DM pixel-level controllability, we employ DINO-V2~\cite{DINOV2} as the feature extractor $\psi$ for garment images $g$ in ControlNet. Unlike CLIP, DINO-V2 not only encodes images into global tokens but also into patch tokens. This approach helps in preserving more pixel information of garment images $g$, offering a detailed representation. Additionally, we implement a fully connected layer (FC) to encode garment features $\psi(g)$ into the space where U-Net resides. Subsequently, these features $\text{FC}(\psi(g))$ are integrated into U-Net through a cross-attention mechanism~\cite{Cross-Attention}. By enhancing GC-DM's feature extraction capabilities for garment images, we improve its controllability at the pixel level.

\begin{figure}
    \centering
    \includegraphics[width=1\linewidth]{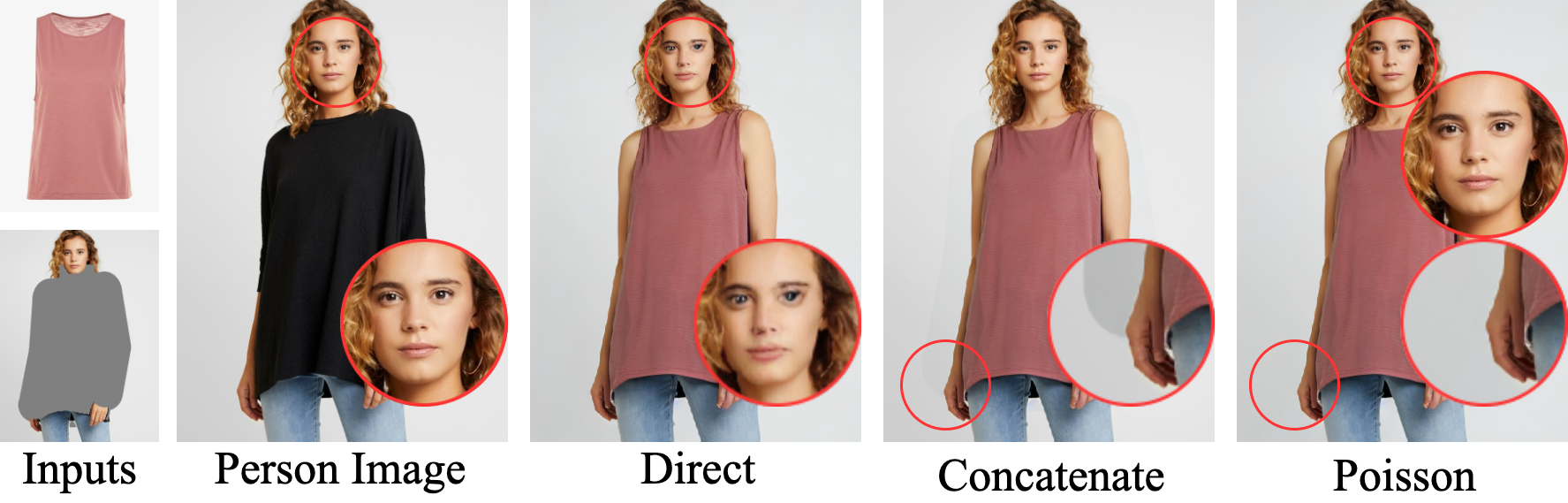}
    \caption{
Results from different types of generation methods. The directly generated try-on images exhibit noticeable distortion in the face region. The results obtained through image concatenating have incongruities at the junctions of the images. This issue is resolved in the results obtained using Poisson blending.}
    \label{Fig:4_Poisson}
\end{figure}

\textbf{Poisson blending.}
PBE \cite{PBE}, as a Latent Diffusion Model (LDM)\cite{LDM}, uses pre-trained autoencoders\cite{AutoencoderKL} to convert images into latent space, thereby reducing computational demands. However, this conversion can cause pixel precision loss during image reconstruction, especially in complex images like faces, leading to noticeable differences from the original, as shown in \cref{Fig:4_Poisson}.

In virtual try-on applications, we typically desire to replace the garment within a designated mask region, while keeping the area outside the mask unchanged. A straightforward approach is to concatenate the input image with the generated image using the mask. However, as shown in \cref{Fig:4_Poisson}, this method can result in noticeable discontinuities at the junction of the two images.

To address the aforementioned issues, we adapt Poisson blending~\cite{Poisson} to seamlessly integrate the input image with the generated image. In particular, given a directly generated virtual try-on image $f^*$, the original person image $h$, and the non-clothing region $\Omega$. The blending image $f$ should satisfy the following equation:

\begin{equation}
    \begin{cases}
        |N_p|f_p-\sum_{q\in N_p}f_q=\sum_{q\in N_p}v_{pq} & p\in\Omega \\
        f_p = f^*_p & p\in\partial\Omega 
    \end{cases}
    \label{eq4}
\end{equation}
where $N_p$ is the set of 4-connected neighbors~\cite{DeepImageBlending} for pixel $p$, $|N_p|$  denotes the number of pixels in the set $N_p$ and $\partial\Omega$ represents of the boundary around $\Omega$. The difference between pixel $p$ and its neighboring pixel $q$ is denoted as $v_{pq}=h_p-h_q$.

As shown in \cref{Fig:4_Poisson}, the adoption of Poisson blending not only ensures that the areas outside the garment region remain unchanged, but also resolves the traces caused by image stitching.

\subsection{Truncation-Based Acceleration Strategy}
\label{subsection:TB-AS}

As shown in \cref{Fig:3_CAT-DM}, for the conventional DDPMs~\cite{DDPM}, when the number of denoising steps $T$ is reduced, the true denoising distribution $q(\mathbf{x}_{t-1}|\mathbf{x}_t)$ is not approximate to a Gaussian distribution and usually intractable. Although denoising diffusion implicit models (DDIMs)~\cite{DDIM} introduce a non-Markovian diffusion process to accelerate sampling, it still requires dozens of samples to generate high-quality images. Hence, significant computational resources are needed even for inference with a trained model, limiting the research and application of diffusion models.
On the other hand, we have observed that diffusion models, when generating patterns or text on garment, exhibit a clear disadvantage compared to GANs.

Also shown in \cref{Fig:3_CAT-DM}, the core idea of TDPM~\cite{TDPM} is to repurpose the diffusion model as an implicit generator to generate the starting point of the reverse diffusion chain. Compared to iteratively denoising starting from Gaussian-distributed noise, the reverse diffusion chain in TDPM is much shorter, i.e., $T_{\text{trunc}}\ll T$. However, the truncated chain of TDPM has an unknown corrupted data distribution at step $T_{\text{trunc}}$, and it is often challenging to learn the implicit distribution $\mathbf{x}_{T_{\text{trunc}}}$ at step $T_{\text{trunc}}$ merely by reusing the parameters of the diffusion model. Besides, TDPM does not hold an advantage in terms of generation quality, especially when faced with garments containing complex patterns or text, i.e., the issue of uncontrollable generation still persists.

\input{table/DressCode}

We introduce a truncation-based acceleration strategy to reduce the sample steps while effectively addressing the issue of uncontrollable generation of complex patterns. As shown in \cref{Fig:3_CAT-DM}, CAT-DM consists of GC-DM and a pre-trained GAN-based virtual try-on model, integrated through the truncation-based acceleration strategy. It utilizes a pre-trained GAN-based model to generate the initial try-on image $\bar{\mathbf{x}}$. We achieve the implicit distribution $\mathbf{x}_{T_{\text{trunc}}}$ at step $T_{\text{trunc}}$ for this image by adding noise through the following equation:

\begin{equation}
    \mathbf{x}_{T_{\text{trunc}}} = \sqrt{\bar{\alpha}_{T_{\text{trunc}}}}\bar{\mathbf{x}}+\sqrt{1-\bar{\alpha}_{T_{\text{trunc}}}}\epsilon, \quad 
 \epsilon\sim\mathcal{N}(0,\mathbf{I})
    \label{eq5}
\end{equation}

In CAT-DM, $\mathbf{x}_{T_{\text{trunc}}}$ serves as the starting point of the reverse diffusion chain, followed by iterative denoising of the noisy image $\mathbf{x}_{T_{\text{trunc}}}$ via GC-DM. Unlike TDPM, we use DDIMs as sampler for generating high-quality samples more rapidly.
The training process is consistent with \cref{eq3}, with the sole difference being that $t$ no longer follows a uniform distribution over 
$\{1,2,...,T\}$, but rather over $\{1,2,...,T_{\text{trunc}}\}$.

By adjusting the size of $T_{\text{trunc}}$ within CAT-DM, we can control the contribution ratio of the pre-trained GAN and GC-DM to the final generated image. Generally speaking, a larger $T_{\text{trunc}}$ results in a greater influence of GC-DM on the final image, while a smaller $T_{\text{trunc}}$ makes the final image lean more towards the result generated by the pre-trained GAN-based model.

%% file: table/DressCode.tex
\begin{table*}
    \centering
    \resizebox{\linewidth}{!}{
        \begin{tabular}{@{}lllllllllllllllllll@{}}
            \toprule
            \multirow{2}{*}{Method} & \multicolumn{6}{c}{DressCode-Upper} & \multicolumn{6}{c}{DressCode-Lower} & \multicolumn{6}{c}{DressCode-Dresses} \\
            \cmidrule(lr){2-7} 
            \cmidrule(lr){8-13}
            \cmidrule(lr){14-19}
            & FID$_\text{u}$ $\downarrow$ & KID$_\text{u}$ $\downarrow$ & FID$_\text{p}$ $\downarrow$ & KID$_\text{p}$ $\downarrow$ & SSIM$_\text{p}$ $\uparrow$ & LPIPS$_\text{p}$ $\downarrow$ 
            & FID$_\text{u}$ $\downarrow$ & KID$_\text{u}$ $\downarrow$ & FID$_\text{p}$ $\downarrow$ & KID$_\text{p}$ $\downarrow$ & SSIM$_\text{p}$ $\uparrow$ & LPIPS$_\text{p}$ $\downarrow$ 
            & FID$_\text{u}$ $\downarrow$ & KID$_\text{u}$ $\downarrow$ & FID$_\text{p}$ $\downarrow$ & KID$_\text{p}$ $\downarrow$ & SSIM$_\text{p}$ $\uparrow$ & LPIPS$_\text{p}$ $\downarrow$ \\
            \midrule
            PBE~\cite{PBE}                  & 20.32 & 7.01 & 18.79 & 6.64 & 0.872 & 0.1209 & 24.95 & 7.36 & 22.44 & 6.78 & 0.804 & 0.2108 & 31.25 & 19.09 & 30.04 & 18.44 & 0.761 & 0.2516 \\
            MGD~\cite{MGD}                  & 17.30 & 5.11 & 15.03 & 5.54 & 0.912 & 0.0624 & 16.76 & 4.04 & 13.67 & 3.79 & 0.893 & 0.0689 & 15.11 & 3.36  & 12.14 & 2.41  & 0.844 & 0.1195 \\
            LaDI-VTON~\cite{LaDI-VTON}      & 17.40 & 5.92 & 14.91 & 6.01 & 0.915 & 0.0649 & 17.90 & 5.45 & 13.76 & 4.61 & \textbf{0.910} & \textbf{0.0596} & 16.13 & 4.76  & 13.00 & 4.05  & 0.854 & \textbf{0.1076} \\
            \midrule
            GC-DM       & \textbf{12.62} & \textbf{1.89} & \textbf{9.85}  & \textbf{2.38} & \textbf{0.927} & \textbf{0.0507} & \textbf{14.83} & \textbf{2.82} & \textbf{10.25} & \textbf{1.81} & 0.902 & 0.0621 & \textbf{14.30} & \textbf{3.36}  & \textbf{10.71} & \textbf{2.02}  & \textbf{0.863} & 0.1091 \\
            \bottomrule
        \end{tabular}
    }

    \caption{Quantitative results on DressCode~\cite{DressCode}. The subscripts `u' and `p' respectively represent the unpaired setting and paired setting.}

    \label{Table:1_DressCode}
\end{table*}

%% file: sec/4_Experiments.tex
\section{Experiments}

\subsection{Experiments Setting}

\parsection{Datasets}
In this work, we focus on evaluating virtual try-on tasks using two popular datasets: DressCode~\cite{DressCode} and VITON-HD~\cite{VITON-HD}. Both datasets contain high-resolution paired images of in-shop garments and their corresponding human models wearing the garments. 
The DressCode dataset includes three categories: upper-body, lower-body, and dresses.
Test experiments are conducted under both paired and unpaired settings. In the paired setting,
the input garment images and the garment worn by the model are the same item. Conversely, in the unpaired setting, a different garment is selected for the virtual try-on task.

\parsection{Evaluation Metrics}
To assess our model quantitatively, we use evaluation metrics to evaluate the coherence and realism of the generated output. In the paired and unpaired settings, we employ the Fréchet Inception Distance (FID)~\cite{FID} and the Kernel Inception Distance (KID)~\cite{KID} to evaluate the realism of the generated output. Furthermore, in the paired setting with available ground truth, we additionally employ the Learned Perceptual Image Patch Similarity (LPIPS)~\cite{LPIPS} and the Structural Similarity Index Measure (SSIM)~\cite{SSIM} to evaluate the coherence of the generated image.

\input{table/VITONHD}

\begin{figure}
    \centering
    \includegraphics[width=1\linewidth]{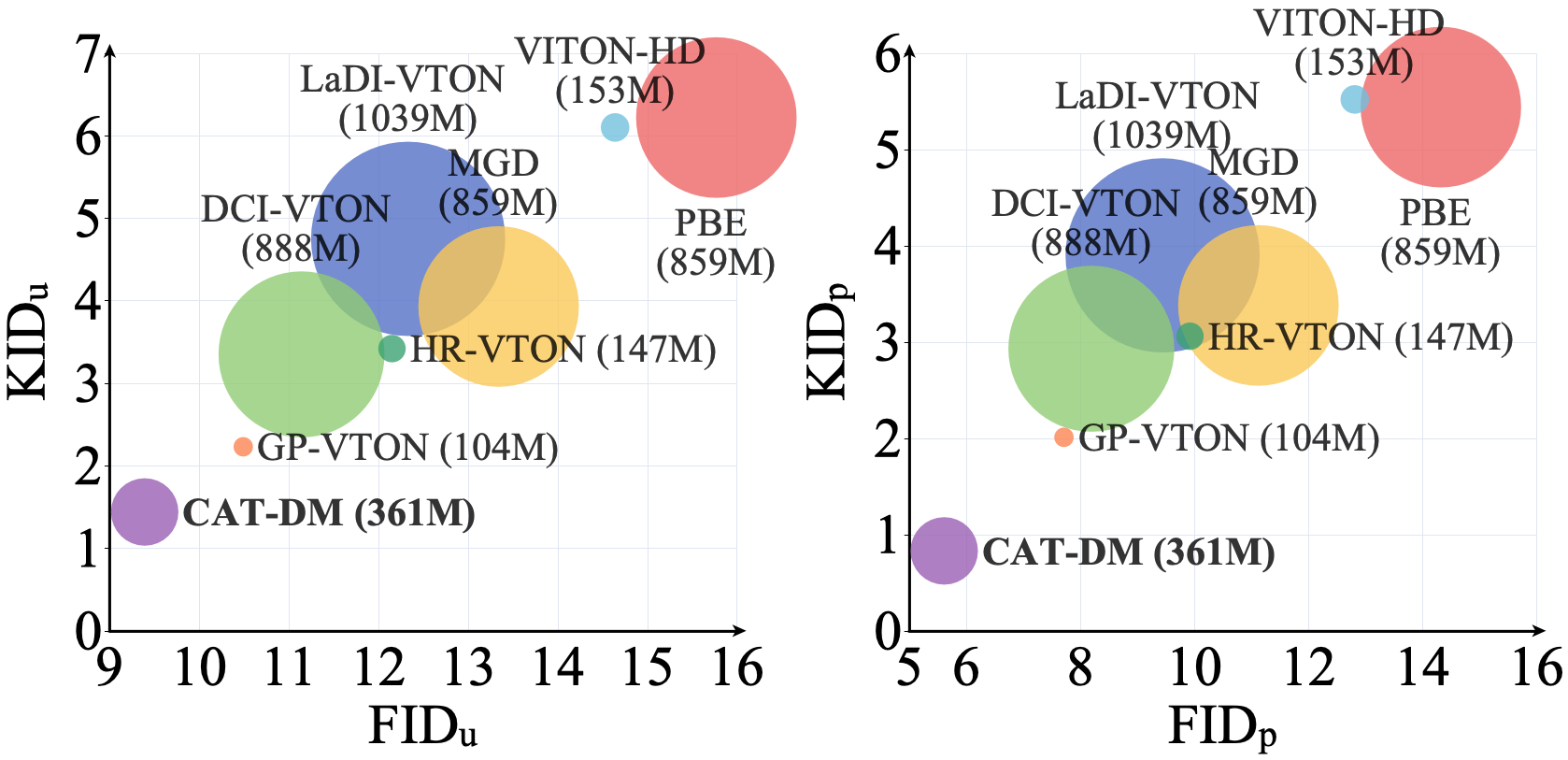}
    \caption{
    Comparative analysis of our method (CAT-DM) with other techniques using the VITON-HD dataset~\cite{VITON-HD}, focusing on the realism of results (better at the bottom left) and the number of trainable parameters (smaller is better). The unpaired setting is on the left, and the paired setting is on the right.
    }
    \label{Fig:5_Parameter}
\end{figure}

\parsection{Implementation Details}
During the experiments, we use an end-to-end training process. All experiments are conducted using two NVIDIA GeForce RTX 4090 GPUs with image resolutions of $512\times384$. We use the AdamW optimizer, set the learning rate to $2\times10^{-5}$.

\begin{figure*}
    \centering
    \includegraphics[width=1\linewidth]{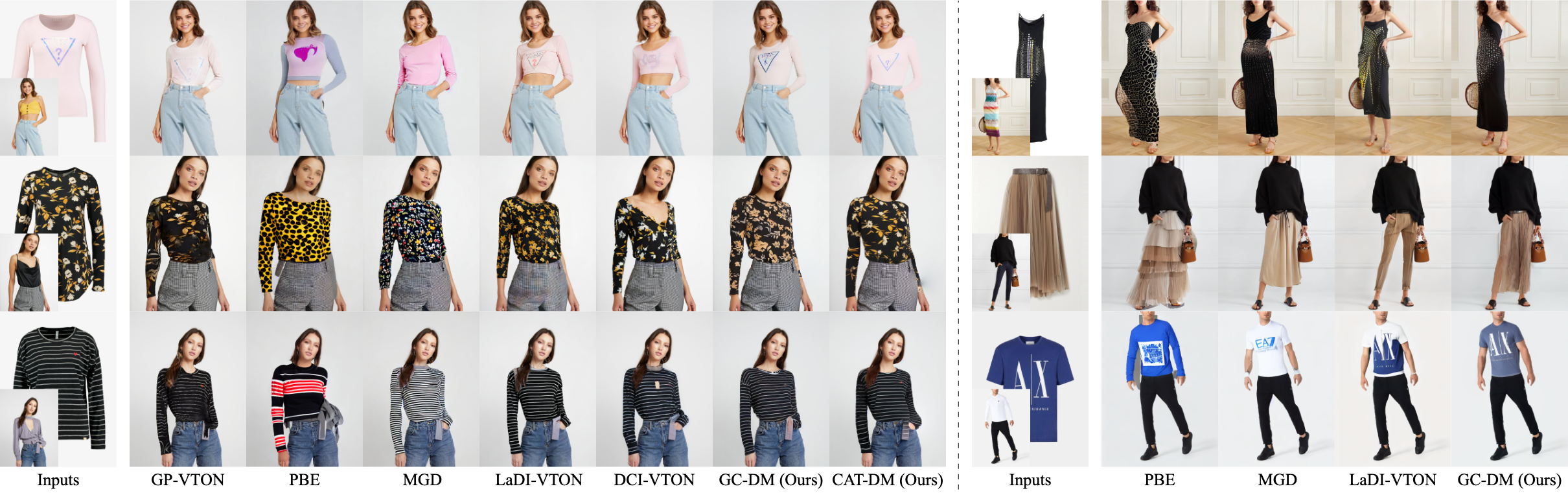}
    \caption{Qualitative results. \textbf{Left}: Comparison results on VITON-HD~\cite{VITON-HD}; \textbf{Right}: Comparison results on DressCode~\cite{DressCode}.}
    \label{Fig:6_Comparison}
\end{figure*}

\subsection{Quantitative Evaluation}

We compare our method with previous virtual try-on methods, including GAN-based virtual try-on models such as VITON-HD~\cite{VITON-HD}, HR-VITON~\cite{HR-VITON}, and GP-VTON~\cite{GP-VTON}, as well as diffusion-based virtual try-on models including MGD~\cite{MGD}, DCI-VTON~\cite{DCI-VTON}, and LaDI-VTON~\cite{LaDI-VTON}. Since our model utilizes PBE~\cite{PBE} as a locked-parameter network, we also include PBE in our comparison. The DressCode dataset was released recently, so we lack pre-trained GAN-based virtual try-on models that have been trained on the DressCode dataset. Therefore, for the quantitative results on the DressCode dataset, we only compare GC-DM instead of CAT-DM with other methods.

We employ two different methods for quantitative comparison of our model. For GC-DM, we use DDIMs sampling with the number of sampling steps set to 16. For CAT-DM, we employ a truncation-based acceleration strategy, utilize a pre-trained GAN-based model with GP-VTON, and set $T_{\text{trunc}}$ to 100 and the number of sampling steps to 2. The generated results for both models are processed using Poisson blending.

As reported in \cref{Table:1_DressCode} and \cref{Table:2_VITON-HD}, our GC-DM outperforms other methods on the majority of metrics, particularly in FID~\cite{FID} and KID~\cite{KID}, demonstrating its effectiveness in image generation quality.  
CAT-DM utilizes a pre-trained GAN as a preconditioner to produce accurate and sharp images. CAT-DM significantly outperforms other methods across all metrics and, notably, can generate ideal images in just two steps, which greatly accelerates the sampling speed of diffusion models. As demonstrated in \cref{Fig:5_Parameter}, compared to other diffusion models, CAT-DM also significantly reduces the number of trainable parameters while also offering a marked advantage in image generation quality.

\subsection{Qualitative Evaluation}

\cref{Fig:6_Comparison} displays the qualitative comparison of GC-DM and CAT-DM with the state-of-the-art baselines on VITON-HD dataset~\cite{VITON-HD} and DressCode dataset~\cite{DressCode} in the unpaired setting, respectively. Based on the test results of PBE, MGD, and LaDI-VTON, it is evident that they struggle to capture the details on the given garment images and cannot accurately reproduce the patterns on the garment. Although DCI-VTON can generate accurate garment patterns to some extent, it fails to detect changes in garment types. This leads to the residual traces of the original garment appearing in the generated virtual try-on images. GP-VTON shows commendable performance in generating accurate images and capturing details, but the resulting images contain some artifacts and lack a degree of realism. Compared to methods based on GANs, frozen CLIP and DINO-V2 benefit from large-scale datasets.

Compared to other diffusion methods, GC-DM shows advantages in both accuracy and realism of generation. The CAT-DM, created by integrating GP-VTON and GC-DM using the truncation-based acceleration strategy, not only rectifies the artifacts present in GP-VTON but also retains the generative capabilities of GC-DM. More importantly, in contrast to other diffusion-based virtual try-on models, CAT-DM can produce high-quality virtual try-on images in just two steps. 

\input{table/Ablation}
\begin{figure}
    \centering
    \includegraphics[width=1\linewidth]{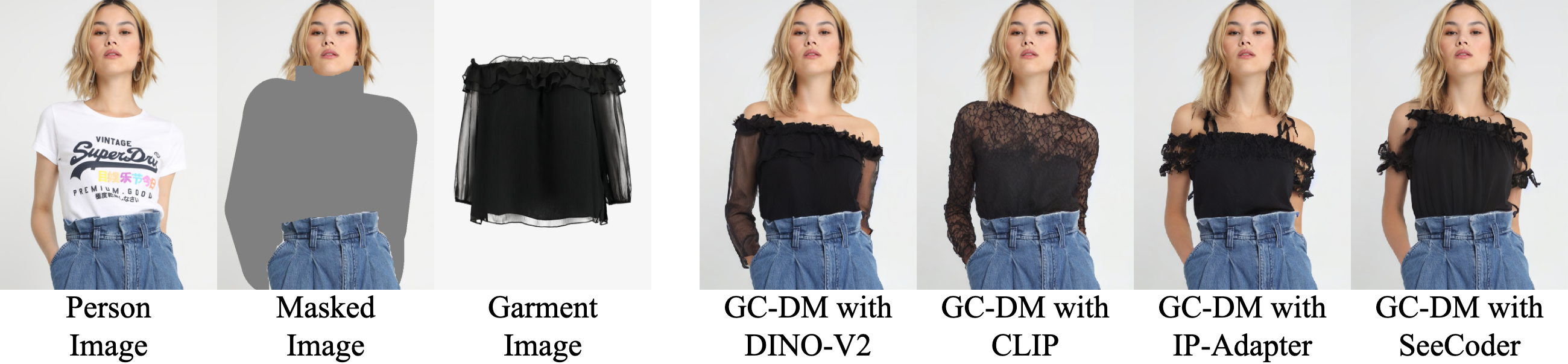}
    \vspace{-5mm}
    \caption{Visual ablation on garment feature extraction.}
    \vspace{-3mm}
    \label{Fig:10_DINO}
\end{figure}

\subsection{Discussion}

\parsection{Garment feature extraction}
We explore the key factors for the garment feature extractor. We compare the results of GC-DM when using CLIP~\cite{CLIP}, DINO-V2~\cite{DINOV2}, IP-Adapter~\cite{IP} and SeeCoder~\cite{SeeCoder} as the garment feature extractor respectively. As reported in \cref{Table:3_Ablation}, for the VITON-HD dataset~\cite{VITON-HD}, with the integration of DINO-V2~\cite{DINOV2}, GC-DM has shown improvement across all metrics. As shown in \cref{Fig:10_DINO}, the GC-DM, utilizing DINO-V2 as a garment feature extractor, is capable of generating more accurate and realistic virtual try-on images. This demonstrates that DINO-V2 can enhance the model's capability to extract features from garment images, thereby also boosting the controllability of the diffusion model. 

\parsection{Poisson blending}
We examine the impact of various processing approaches on the quality of the generated images. As reported in \cref{Table:3_Ablation}, for the VITON-HD dataset~\cite{VITON-HD}, compared to using the frozen encoder-decoder of LDMs~\cite{LDM} to generate virtual try-on images directly, concatenating together the input person image with the generated try-on image can indeed improve the quality of the resulting image. However, the seams at the point of stitching can still impact the image quality. Employing Poisson blending can eliminate such issues, resulting in more realistic virtual try-on images. 

\begin{figure}
    \centering
    \includegraphics[width=1\linewidth]{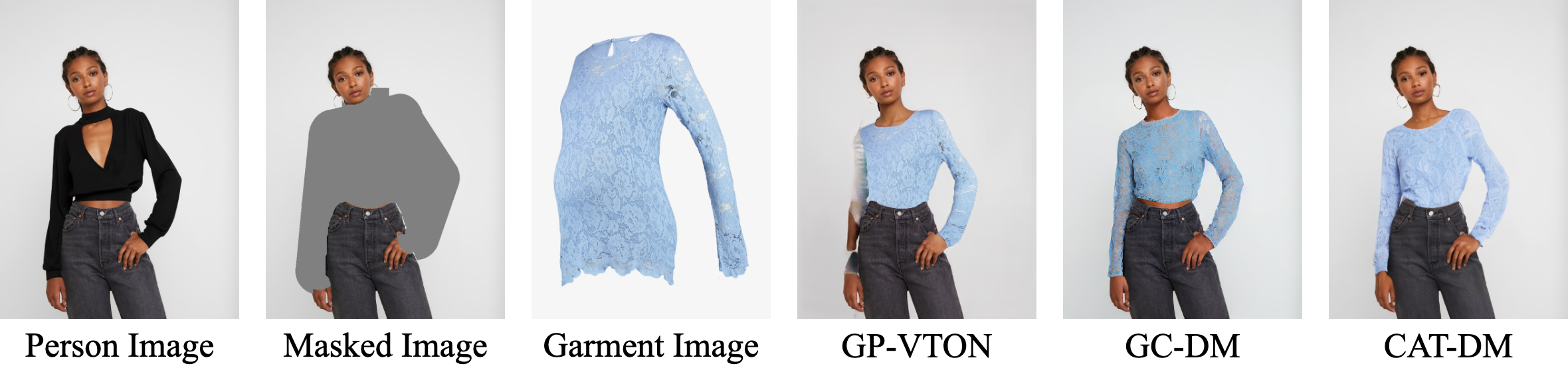}
    \vspace{-3mm}
    \caption{CAT-DM can refine and adjust the try-on results gener- ated by pre-trained GAN-based methods.}
    \label{Fig:8_Refine}

\end{figure}

\parsection{Refinement function of the diffusion model}
The diffusion model can be taken as a refined module. When the pre-trained GAN-based method generate a try-on result with over-distorted warped garment, diffusion model can adjust it. As shown in  \cref{Fig:8_Refine}, the try-on images generated by GP-VTON lack an arm on one side, but CAT-DM is capable of rectifying it.

\begin{figure}
    \centering
    \includegraphics[width=1\linewidth]{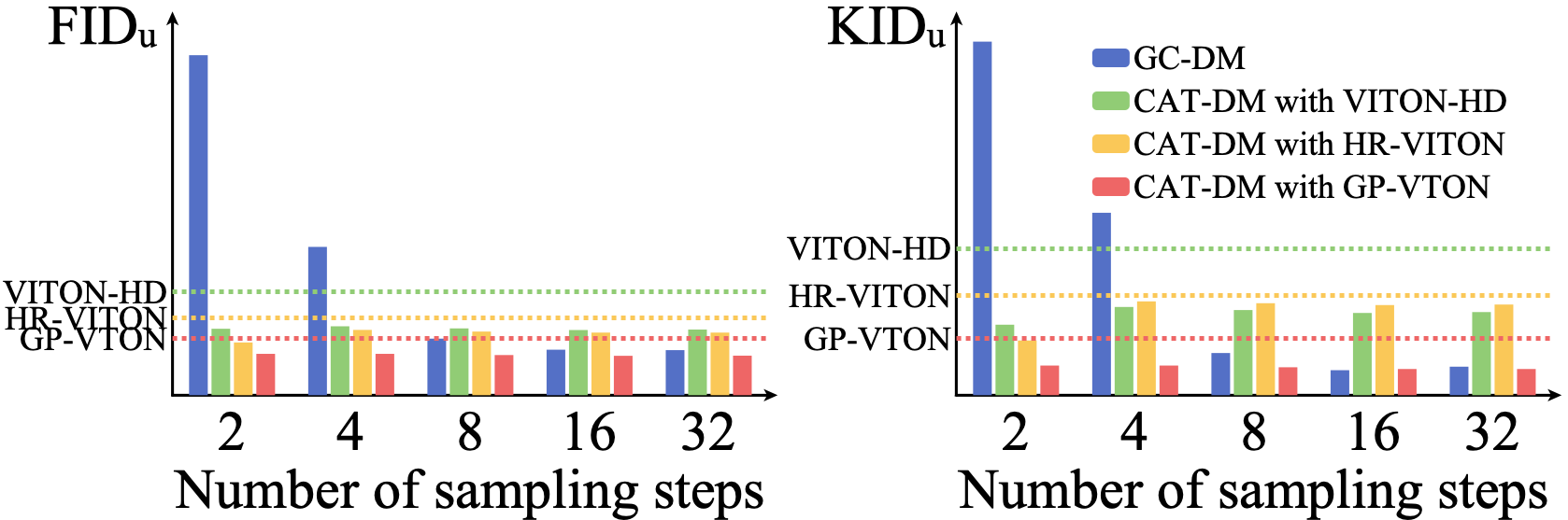}
    \vspace{-3mm}
    \caption{Discussion about the pre-trained GAN-based model. The bar chart represents the performance of GC-DM and CAT-DM using different GAN-based models across various sampling step counts, while the dashed lines indicate the performance of the different GAN-based models.}
    \label{Fig:7_Table_GANs}

\end{figure}

\parsection{Pre-trained GAN-based model}
We explore the impact of different GAN-based models on CAT-DM's performance and compare it with GC-DM, which does not use GAN-based models. The experimental results are shown in \cref{Fig:7_Table_GANs}. For the VITON-HD dataset~\cite{VITON-HD}, the three dashed lines respectively represent the original performance of the three GAN-based methods: VITON-HD, HR-VITON, and GP-VTON. 
Although the performance of GC-DM surpasses that of all GAN-based models when the number of sampling steps is sufficient, the performance of GC-DM significantly degrades when the number of sampling steps is inadequate. 
CAT-DM leverages the rapid generation capabilities of GANs to significantly reduce the need for numerous sampling steps. Compared to GC-DM, CAT-DM avoids performance degradation when the number of sampling steps is low. Additionally, CAT-DM achieves higher performance compared to the GAN-based models it utilizes. 
Furthermore, we note that the performance of CAT-DM is, to a certain extent, reliant on the performance of GAN-based models. 
When the number of sampling steps is sufficient, CAT-DM, utilizing GP-VTON as its pre-trained GAN-based model, not only surpasses GP-VTON but also outperforms GC-DM.

\begin{figure}
    \centering
    \includegraphics[width=1\linewidth]{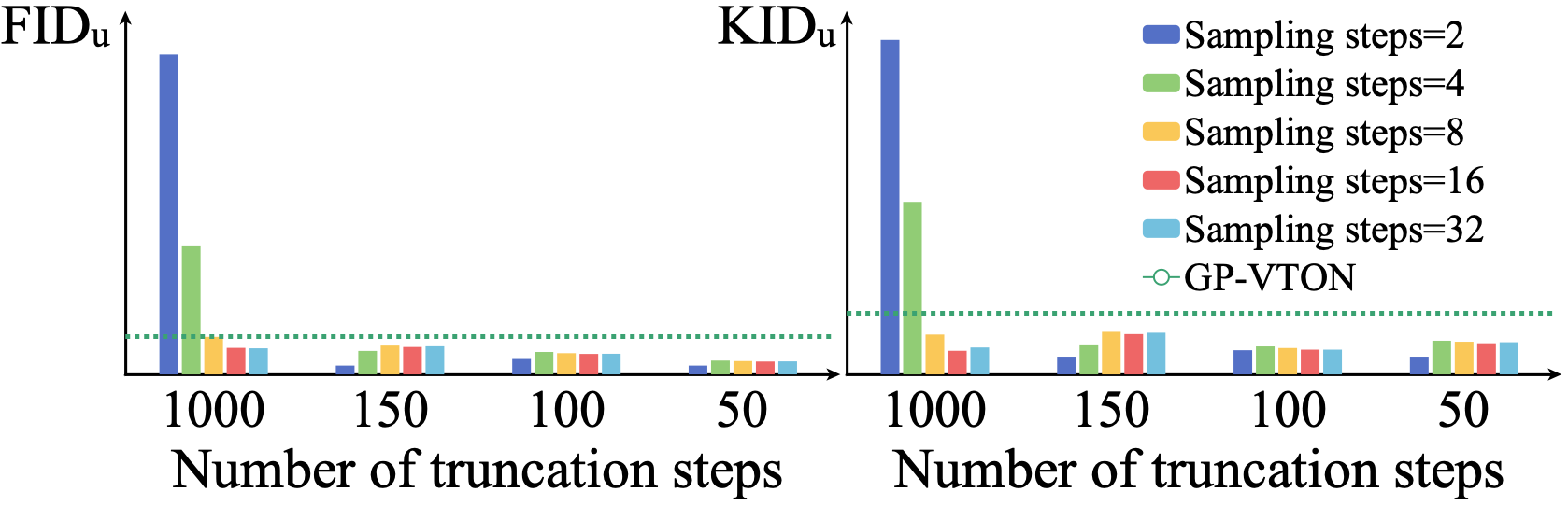}
    \vspace{-3mm}
    \caption{Discussion about the truncation step. The bar chart represents the performance of GC-DM and CAT-DM using different truncation steps across various sampling step counts.}
    \label{Fig:9_Table_Trunc}

\end{figure}

\parsection{Truncation step}
As shown in \cref{Fig:9_Table_Trunc}, we conducte experiments with different truncation settings of $T_\text{Trunc}=$ 0, 50, 100, 150 and 1000. 
On one hand, when $T_\text{Trunc}$ is set to 0, CAT-DM and GP-VTON are essentially the same model. On the other hand, when $T_\text{Trunc}$ is set to 1000, CAT-DM and GC-DM become identical models. 
When $T_\text{Trunc}$ is set to 50, 100, and 150, we observe that the model tends to perform best when the number of sampling steps is 2. 

%% file: table/VITONHD.tex
\begin{table}
    \centering
    \resizebox{\linewidth}{!}{
        \begin{tabular}{@{}lllllll@{}}
            \toprule
            Method & FID$_\text{u}$ $\downarrow$ & KID$_\text{u}$ $\downarrow$ & FID$_\text{p}$ $\downarrow$ & KID$_\text{p}$ $\downarrow$ & SSIM$_\text{p}$ $\uparrow$ & LPIPS$_\text{p}$ $\downarrow$ \\
            \midrule
            VITON-HD~\cite{VITON-HD}        & 14.64 & 6.10 & 12.81 & 5.52 & 0.848 & 0.1216  \\
            HR-VITON~\cite{HR-VITON}        & 12.15 & 3.42 & 9.92  & 3.06 & 0.860 & 0.1038  \\
            GP-VTON~\cite{GP-VTON}          & 10.49 & 2.23 & 7.71  & 2.01 & 0.857 & 0.0897  \\
            \midrule     
            PBE~\cite{PBE}                  & 15.77 & 6.22 & 14.32 & 5.44 & 0.763 & 0.2254  \\
            MGD~\cite{MGD}                  & 13.34 & 3.93 & 11.12 & 3.38 & 0.827 & 0.1280  \\
            LaDI-VTON~\cite{LaDI-VTON}      & 12.33 & 4.75 & 9.44  & 3.90 & 0.861 & 0.0968  \\
            DCI-VTON~\cite{DCI-VTON}        & 11.14 & 3.35 & 8.19  & 2.93 & 0.875 & 0.0816  \\
            \midrule 
            GC-DM       & 9.67  & \textbf{1.36} & 7.11  & 1.12 & 0.862 & 0.0988  \\
            CAT-DM      & \textbf{8.93}  & 1.37 & \textbf{5.60}  & \textbf{0.83} & \textbf{0.877} & \textbf{0.0803}  \\
            \bottomrule
        \end{tabular}
    }
    \caption{Quantitative results on VITON-HD~\cite{VITON-HD}. The subscripts `u' and `p' respectively represent the unpaired setting and paired setting.}
    \label{Table:2_VITON-HD}
\end{table}

%% file: table/Ablation.tex
\begin{table}
    \centering
    \resizebox{\linewidth}{!}{
        \begin{tabular}{@{}llllllll@{}}
            \toprule
            Extractor & Process & FID$_\text{u}$ $\downarrow$ & KID$_\text{u}$ $\downarrow$ & FID$_\text{p}$ $\downarrow$ & KID$_\text{p}$ $\downarrow$ & SSIM$_\text{p}$ $\uparrow$ & LPIPS$_\text{p}$ $\downarrow$ \\
            \midrule
            \multirow{3}{*}{DINO-V2~\cite{DINOV2}}              & Direct Generation  & 10.76 & 2.53 & 8.25 & 2.09 & 0.835 & 0.1069  \\
                                                                & Concatenation      & 10.57 & 2.59 & 8.18 & 2.42 & 0.854 & 0.1033  \\
                                                                & Poisson Blending   & \textbf{9.67}  & \textbf{1.36} & \textbf{7.11} & \textbf{1.12} & \textbf{0.862} & \textbf{0.0988}  \\
            \midrule
            CLIP~\cite{CLIP}                                    & Poisson Blending   & 10.21 & 1.77 & 7.90 & 1.38 & 0.853 & 0.1111  \\
            \midrule
            IP-Adapter~\cite{IP}                        & Poisson Blending   & 11.23 & 3.90 & 8.13 & 2.86 & 0.847 & 0.1127  \\
            \midrule
            SeeCoder~\cite{SeeCoder}                            & Poisson Blending   & 9.94  & 1.66 & 7.13 & 1.58 & 0.856 & 0.1049  \\
            \bottomrule
        \end{tabular}
    }
    \caption{Discussion about extractors and Poisson blending. }
    \label{Table:3_Ablation}
\end{table}

%% file: sec/5_Conclusion.tex
\section{Conclusion}
To enhance the controllability of diffusion models in virtual try-on tasks and accelerate the sampling speed of these models, we introduce the CAT-DM. It combines a specially designed try-on model, GC-DM, with a pre-trained GAN model, utilizing an innovative truncation-based acceleration strategy. Specifically, to enhance the generation of detailed garment textures, GC-DM improves the feature extraction from garment images. Additionally, by adopting the ControlNet architecture, GC-DM introduces extra control conditions, thereby increasing the controllability of the diffusion model. To accelerate the sampling speed of diffusion models, CAT-DM initiates a reverse denoising process with an implicit distribution generated by a pre-trained GAN-based model. A substantial number of experiments demonstrate the superiority of our method in terms of image quality, controllability, and sampling speed. The limitation of our CAT-DM will be discussed in the supplement.